%% file: main.tex
\documentclass[10pt,twocolumn,letterpaper]{article}

\usepackage{cvpr}              % To produce the CAMERA-READY version
\usepackage{multirow}
\usepackage{hyperref}
\usepackage[accsupp]{axessibility} 
\definecolor{cvprblue}{rgb}{0.21,0.49,0.74}
\def\paperID{} % *** Enter the Paper ID here
\def\confName{}
\def\confYear{}

\title{Cross-modal Fuzzy Alignment Network for Text-Aerial Person Retrieval and A Large-scale Benchmark}

\author{Yifei Deng\textsuperscript{1,2}, Chenglong Li\textsuperscript{1,}\textsuperscript{*}, Yuyang Zhang\textsuperscript{4}, Guyue Hu\textsuperscript{3}, Jin Tang\textsuperscript{2}\\
\textsuperscript{1}State Key Laboratory of Opto-Electronic Information
Acquisition and Protection Technology\\
\textsuperscript{2}School of
Computer Science and Technology, Anhui University\\ 
\textsuperscript{3}School of Artificial
Intelligence, Anhui University        \textsuperscript{4}The University of Hong Kong\\
{\tt\small \{yf-ah, lcl1314\}@foxmail.com  u3604018@connect.hku.hk  \{guyue.hu, tangjin\}@ahu.edu.cn}
}

\begin{document}
\maketitle

\renewcommand\thefootnote{\textasteriskcentered}
\footnotetext{Corresponding author.}

\input{sec/0_abstract}    
\input{sec/1_intro}
\input{sec/2_formatting}

\input{sec/3_finalcopy}
{
    \small
    \bibliographystyle{ieeenat_fullname}
    \bibliography{main}
}


\end{document}

%% file: sec/0_abstract.tex
\begin{abstract}
Text-aerial person retrieval aims to identify targets in UAV-captured images from eyewitness descriptions, supporting intelligent transportation and public security applications.
Compared to ground-view text–image person retrieval, UAV-captured images often suffer from degraded visual information due to drastic variations in viewing angles and flight altitudes, making semantic alignment with textual descriptions very challenging. 
To address this issue, we propose a novel Cross-modal Fuzzy Alignment Network, which quantifies the token-level reliability by fuzzy logic to achieve accurate fine-grained alignment and incorporates ground-view images as a bridge agent to further mitigate the gap between aerial images and text descriptions, for text–aerial person retrieval.
In particular, we design the Fuzzy Token Alignment  module that employs the fuzzy membership function to dynamically model token-level association strength and suppress the influence of unobservable or noisy tokens. It can alleviate the semantic inconsistencies caused by missing visual cues and significantly enhance the robustness of token-level semantic alignment.
Moreover, to further mitigate the gap between aerial images and text descriptions, we design a Context-Aware Dynamic Alignment module to incorporates the ground-view agent as a bridge in text–aerial alignment and adaptively combine direct alignment and agent-assisted alignment to improve the robustness.
In addition, we construct a large-scale benchmark dataset called AERI-PEDES by using a chain-of-thought to decompose text generation into attribute parsing, initial captioning, and refinement, thus boosting textual accuracy and semantic consistency.
Experiments on AERI-PEDES and TBAPR demonstrate the superiority of our method.
The dataset and code are available at: https://github.com/Yifei-AHU/AERI-PEDES

\end{abstract}

%% file: sec/1_intro.tex
\section{Introduction}
\label{sec:intro}
\begin{figure}[!t]
\centering
\includegraphics[width=3.2in,height=3.4in]{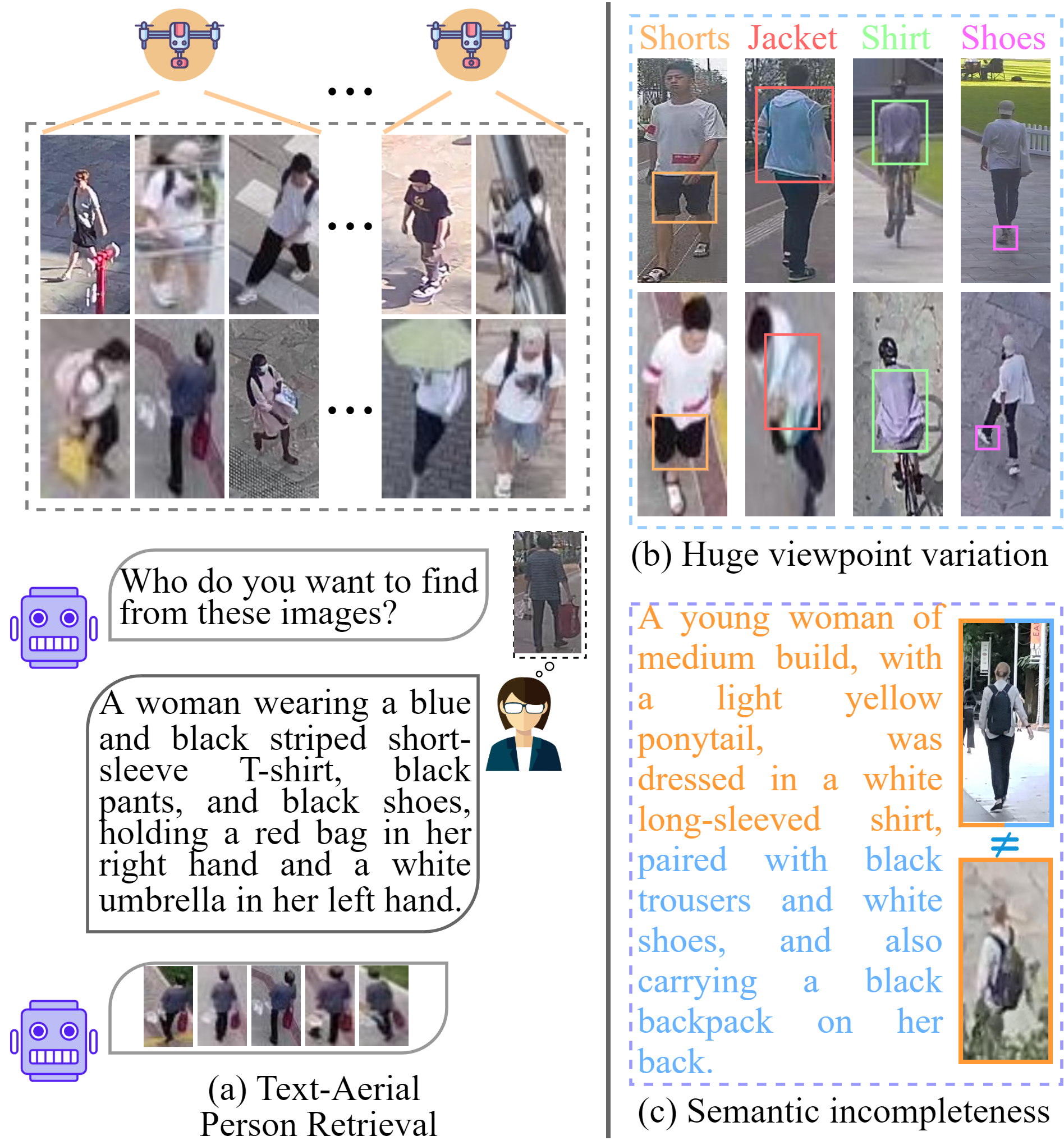}
\caption{(a) Illustration of the text-aerial person retrieval task. (b) Semantically identical targets with substantial viewpoint discrepancies between ground and aerial perspectives. (c) Incomplete visual cues in aerial images result in only partial alignment with the corresponding captions.}
\label{fig_1}
\end{figure}
With the rapid advancement of deep learning~\cite{ZhaXia_Learning_MICCAI2025}~\cite{li2025ckdf}~\cite{li2025enhance}~\cite{pan2025diverse}~\cite{ding2025quality}~\cite{liu2024rgbt} and vision–language models~\cite{shen2025fine}~\cite{li2026toward}~\cite{xu2025stare}~\cite{MIR-2022-06-193}~\cite{MIR-2022-05-167}, Text-Image Person Retrieval (TIPR) has attracted increasing attention in recent years.
TIPR aims to accurately retrieve target individuals from image databases based on captions, with important applications in intelligent surveillance and traffic management~\cite{eom2019learning}~\cite{jin2025sequencepar}~\cite{jin2025pedestrian}.
However, all existing TIPR studies~\cite{jing2020pose}~\cite{deng2025uncertainty}~\cite{li2017person}~\cite{qin2024noisy} rely on data captured by fixed ground-based cameras, which limits the coverage of person observation and makes it difficult to capture dynamic scenes in complex environments.
The widespread adoption of unmanned aerial vehicles (UAVs) offers unique advantages for person monitoring. 
UAVs can capture scenes dynamically from multiple angles and locations, covering areas inaccessible to ground cameras and providing richer information sources.
Building on the flexibility and wide-area coverage of UAVs, extending the TIPR task to aerial image scenarios captured by UAVs holds significant research value and practical importance. 
This direction not only helps address the challenges of cross-modal matching and semantic alignment in complex environments but also fully unleashes the potential of UAVs in applications such as intelligent traffic management, public safety, and security surveillance.
To advance this line of research, Wang et al.~\cite{wang2025aea} are the first to explore the problem of person retrieval from aerial perspectives and proposed the Text-Aerial Person Retrieval (TAPR) task, as shown in Figure~\ref{fig_1}(a). 

Different from conventional TIPR, which only needs to address cross-modal semantic alignment between text and images, person images captured from aerial views exhibit nonlinear distortions in appearance, body posture, and geometric proportions due to extreme variations in shooting angle and altitude, as illustrated in Figure.~\ref{fig_1}(b). This significantly increases the difficulty of aligning them with textual descriptions.
Moreover, query texts are typically derived from eyewitness descriptions and contain more complete and fine-grained person attributes and appearance details. However, in drone-captured scenarios, visual cues of persons are often sparse or even partially missing due to factors such as altitude, viewpoint deviation, and occlusion. As shown in Figure.~\ref{fig_1}(c), persons in aerial views only cover part of the semantic regions described in the text (highlighted in orange), whereas persons in ground views can fully correspond to the textual description.
This inconsistency in visibility can lead to missing effective visual correspondences for some text tokens when constructing token-level fine-grained associations, thereby introducing erroneous cross-modal alignments.

To address the aforementioned challenges, we propose a Cross-Modal Fuzzy Alignment Network (CFAN), which leverages fuzzy logic to quantify token-level reliability for fine-grained alignment and incorporates ground-view images as a bridging proxy to further reduce the gap between aerial images and captions. 
Specifically, to mitigate the impact of visual discrepancies caused by aerial images captured at different altitudes on cross-modal alignment, we design a Context-Aware Dynamic Alignment (CDA) module. 
By using ground images as a bridge, this module quantifies the alignment difficulty between text–aerial and text–ground images, and dynamically adjusts the contributions of direct alignment and bridged alignment, thereby enhancing the stability of text–aerial image alignment.
Moreover, to address semantic inconsistencies in fine-grained alignment caused by missing visual cues, we propose a Fuzzy Token Alignment (FTA) module.
This module dynamically models token-level association strength via fuzzy membership functions while suppressing unobservable or noisy tokens, significantly improving the robustness of token-level semantic alignment.

To further advance research on TAPR task, we construct a large-scale benchmark, AERI-PEDES, which contains 112,672 person images captured from diverse cameras and varied scenarios. 
For caption, to reduce manual annotation costs, we design a Chain-of-Thought (CoT)~\cite{wei2022chain} based generation framework that decomposes the caption generation task into structured reasoning steps using multimodal language models, ensuring the generated training captions are rich in fine-grained attributes and visually consistent. 
To better reflect real-world applications, the test captions are manually annotated to ensure natural semantic expression, allowing the evaluation results to accurately reflect the true performance of the models. 
Extensive experiments on two text-aerial person retrieval benchmarks have thoroughly validated the effectiveness of the proposed method.
In summary, the main contributions of this paper are as follows:
\begin{itemize}
\item{We propose a Cross-modal Fuzzy Alignment Network, which quantifies the token-level reliability by fuzzy logic to achieve accurate fine-grained alignment, and incorporates ground-view images as a bridge agent to further mitigate the gap of text-aerial image.}
\item{We design a Context-Aware Dynamic Alignment module, which incorporates ground images as a bridge and quantifies alignment difficulty to adaptively balance direct and bridged alignment, achieving robust cross-modal alignment.}
\item{We introduce the Fuzzy Token Alignment module, which models token-level reliability with a fuzzy membership function, enhancing shared token alignment while suppressing non-shared alignment, leading to improved fine-grained alignment.}
\item{We construct a large-scale benchmark, AERI-PEDES, and develop a Chain-of-Thought based caption generation framework to produce fine-grained, visually consistent captions.}
\end{itemize}

%% file: sec/2_formatting.tex
\section{Related Work}
\subsection{Text-Image Person Retrieval}
TIPR aims to accurately locate corresponding person images from a gallery based on natural language descriptions. The task is first introduced by Li et al.~\cite{li2017person}, who establish the CUHK-PEDES dataset, laying the foundation for cross-modal person retrieval research. Subsequently, extended benchmarks such as ICFG-PEDES~\cite{zhu2021dssl} and RSTPReid~\cite{ding2021semantically} are released, further advancing TIPR toward more realistic scenarios.
Early studies mainly rely on global feature alignment~\cite{zheng2020dual}, which reduces the modality gap between vision and language by constructing a joint embedding space. However, these approaches typically focus on salient regions, making it difficult to capture fine-grained cross-modal semantic correspondences. 
To address this, later works gradually shift toward local feature matching. For instance, Chen et al.~\cite{chen2022tipcb} capture richer semantic information through multi-scale features to achieve adaptive cross-modal alignment, while Wang et al.~\cite{wang2020vitaa} explore the association between human attributes and text, effectively improving fine-grained alignment accuracy.
In recent years, the development of Vision-Language Pre-trained Models~\cite{radford2021learning}~\cite{li2023blip}~\cite{li2021align} has further driven TIPR research. Methods based on large-scale pre-trained models such as CLIP explore multi-layer semantic mining~\cite{yan2023clip}, image-guided language reconstruction~\cite{jiang2023cross}, and hierarchical supervision from unimodal teachers~\cite{deng2025learning}, significantly enhancing cross-modal alignment accuracy and generalization. Moreover, pre-training with large-scale image-text data generated by diffusion models or VLMs~\cite{tan2024harnessing}~\cite{yang2023towards}~\cite{jiang2025modeling} also substantially improves retrieval robustness.

\subsection{Fuzzy Deep Learning}
Fuzzy logic, originally proposed by Lotfi Zadeh~\cite{zadeh1965fuzzy}, aims to characterize the inherent uncertainty and ambiguity of the real world.
In recent years, researchers have integrated fuzzy logic with neural networks, forming the paradigm of Fuzzy Deep Learning. 
This integration endows models with the ability to reason under uncertainty and ambiguity, demonstrating remarkable advantages in fields such as medical image analysis~\cite{ieracitano2022fuzzy}~\cite{golcuk2022hybrid} and cross-modal retrieval~\cite{duan2025fuzzy}.
Compared with traditional deep learning, fuzzy deep learning excels at modeling fuzzy boundaries and quantifying model uncertainty. 
For example, Ding et al.~\cite{ding2021multimodal} introduced fuzzy adjacency information to optimize MRI brain structure segmentation.
Akram et al.~\cite{ahmadi2021fwnnet} proposed bipolar fuzzy neurons to achieve more accurate medical diagnoses; 
Nan et al.~\cite{nan2023fuzzy} incorporated fuzzy logic into attention mechanisms to construct a fuzzy attention network for airway segmentation, significantly improving segmentation performance.
In the field of cross-modal retrieval, Duan et al.~\cite{duan2025fuzzy} proposed the Fuzzy Multimodal Learning framework based on fuzzy set theory, achieving more reliable retrieval through self-estimated cognitive uncertainty. 
Xu et al.~\cite{xu2025residual} introduced a hypergraph-based residual fuzzy alignment network for open-set 3D cross-modal retrieval, significantly improving generalization. 

\section{Method}
\begin{figure*}[!t]
\centering
\includegraphics[width=6.8in,height=3.2in]{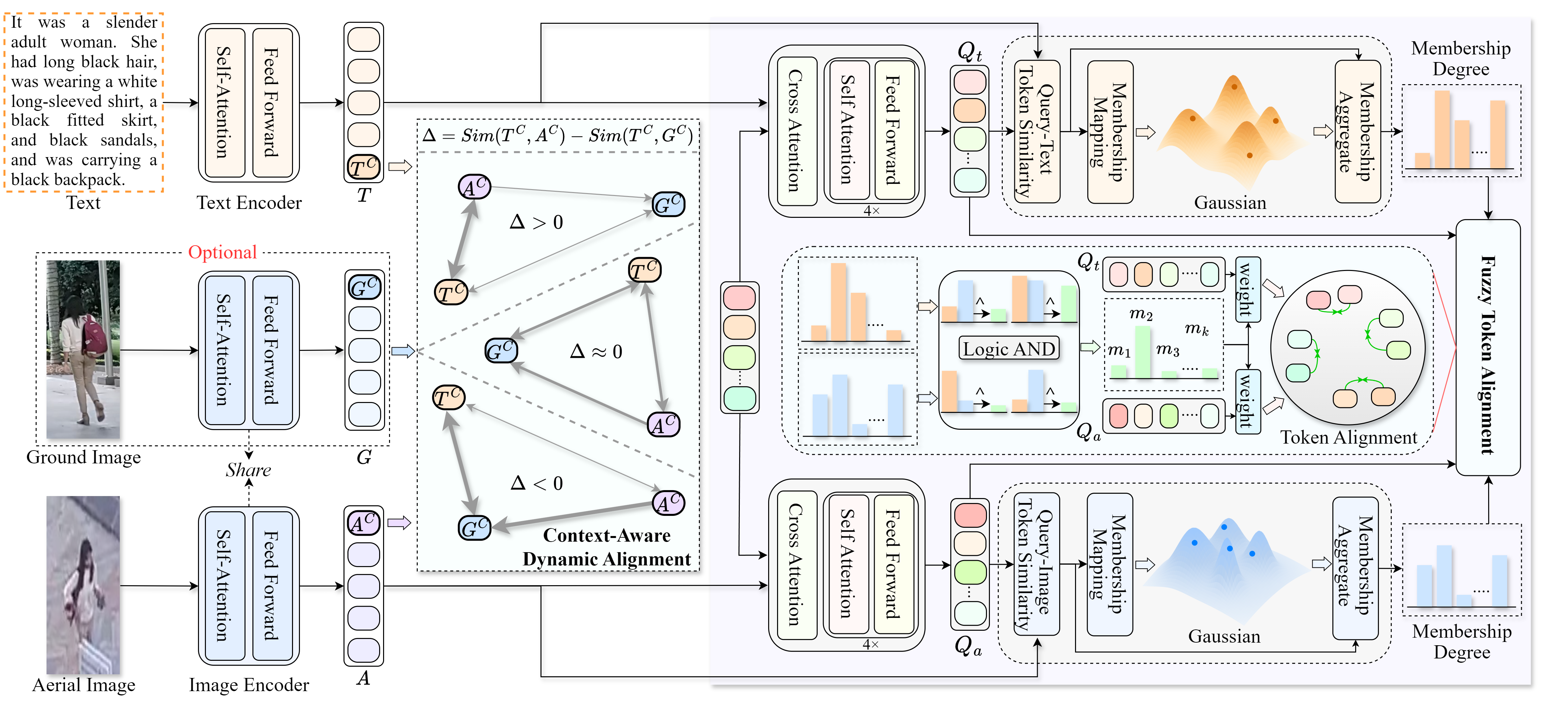}
\caption{Overview of the proposed Cross-Modal Fuzzy Alignment Network.}
\label{fig_2}
\end{figure*}
\subsection{Overview}
Figure~\ref{fig_2} illustrates our proposed Cross-Modal Fuzzy Alignment Network, which consists of two modules, the Context-Aware Dynamic Alignment (CDA) module and the Fuzzy Token Alignment (FTA) module. 
We employ a shared CLIP image encoder to extract aerial person image features $A$ and ground-level person image features $G$, and a CLIP text encoder to extract person description features $T$.
The CDA module takes the global representations of aerial features $A^c$, ground features $G^c$, and text features $T^c$ as input.
It evaluates the alignment difficulty of each sample by comparing text-aerial and text-ground similarities, and dynamically modulates the contributions of direct and bridge-mediated alignment, thereby achieving robust sample-level cross-modal alignment.
The FTA module takes $A$ and $T$ as input and introduces a fuzzy membership function at the token level to assign each token a continuous existence degree. 
Multi-source membership degrees are fused via a logical AND operation to semantically modulate token-level alignment, effectively suppressing interference from unobservable or noisy tokens and achieving fine-grained, robust cross-modal token-level alignment.
By combining CDA and FTA, our framework simultaneously achieves adaptive sample-level alignment and robust fine-grained token-level alignment, effectively bridging the semantic gap between text and aerial images.

\subsection{Context-Aware Dynamic Alignment}
We propose a Context-Aware Dynamic Alignment (CDA) module to tackle the cross-modal alignment challenges caused by the visual discrepancies in aerial images. 
This module leverages ground-view images as auxiliary guidance, and quantify the alignment difficulty of text–ground and text–aerial, dynamically adjusts the weighting between direct alignment and ground-assisted bridged alignment. Specifically, given the global text features $T^C \in \mathbb{R}^{B \times D}$, global aerial image features $A^C \in \mathbb{R}^{B \times D}$, and global ground image features $G^C \in \mathbb{R}^{B \times D}$, we first compute the cross-modal similarity differences for each sample:
\begin{equation}
\Delta_i = \text{sim}(\mathbf{T_i^C}, \mathbf{A_i^C}) - \text{sim}(\mathbf{T_i^C}, \mathbf{G_i^C}), \quad i = 1, \dots, B,
\end{equation}
where $\text{sim}(\cdot, \cdot)$ denotes the cosine similarity. 
Intuitively, $\Delta_i > 0$ indicates that direct alignment is likely sufficient, whereas $\Delta_i < 0$ implies that a bridge via ground-level images is necessary.
To enable dynamic alignment, we map the similarity discrepancy to a continuous coefficient $\alpha_i \in [0,1]$ using a nonlinear, context-sensitive activation:
\begin{equation}
\alpha_i = \frac{1}{1 + \exp\Big[-k \cdot \Delta_i]},
\end{equation}
where $k$ is a sensitivity hyperparameter controlling the steepness of the mapping. 
Conceptually, $\alpha_i$ functions as a soft decision gate, where samples with high direct similarity $(\text{sim}(\mathbf{T_i^C}, \mathbf{A_i^C}) \gg \text{sim}(\mathbf{T_i^C}, \mathbf{G_i^C}))$ yield $\alpha_i \to 1$, emphasizing direct text-aerial alignment, whereas samples with low direct similarity $(\text{sim}(\mathbf{T_i^C}, \mathbf{A_i^C}) \ll \text{sim}(\mathbf{T_i^C}, \mathbf{G_i^C}))$ yield $\alpha_i \to 0$, emphasizing bridge-based alignment via ground images.
Therefore, the loss of this module is defined as:
\begin{equation}
\begin{split}
\mathcal{L}_{\text{CAD}} = \frac{1}{B} \sum_{i=1}^{B} \Big[ \alpha_i \cdot \mathcal{L}_{\text{direct}}(\mathbf{T_i^C}, \mathbf{A_i^C}) + \\
(1 - \alpha_i) \cdot \mathcal{L}_{\text{bridge}}(\mathbf{T_i^C}, \mathbf{G_i^C}, \mathbf{A_i^C}) \Big],
\end{split}
\end{equation}
where $\mathcal{L}_{\text{direct}}$ denotes the direct alignment loss between text and aerial image, and $\mathcal{L}_{\text{bridge}}$ represents the indirect alignment of text and aerial features via a ground-level image as a semantic bridge. 
Both alignment terms are implemented using the SDM loss~\cite{jiang2023cross}:
\begin{equation}
\mathcal{L}_{\text{direct}}(\mathbf{T_i^C}, 
\mathbf{A_i^C}) = \text{SDM}(\mathbf{T_i^C}, \mathbf{A_i^C}),
\end{equation}
\begin{equation}
\begin{split}
\mathcal{L}_{\text{bridge}}(\mathbf{T_i^C}, \mathbf{G_i^C}, \mathbf{A_i^C}) = \text{SDM}(\mathbf{T_i^C}, \mathbf{G_i^C}) + \\ 
\text{SDM}(sg(\mathbf{G_i^C}), \mathbf{A_i^C}),
\end{split}
\end{equation}
where $sg(·)$ means stop-gradient operator, which prevents backpropagation through the ground-level features, ensuring that the aerial features are adaptively aligned toward the ground bridge without altering the ground representation.
This design enables each sample to dynamically balance direct and bridge-mediated alignment, effectively mitigating cross-modal alignment challenges caused by variations in visual cues. Moreover, when ground images are unavailable, the CDA loss degenerates to the standard SDM loss.

\subsection{Fuzzy Token Alignment}
To address fine-grained misalignment between text and aerial person images, we propose the Fuzzy Token Alignment (FTA) module, grounded in fuzzy logic theory to quantify token reliability within each modality. 
Tokens with low membership are considered unreliable and their contribution to cross-modal alignment is suppressed, whereas tokens with high membership in both modalities are preserved, enabling precise alignment.

Given textual features $\mathbf{T} \in \mathbb{R}^{B \times N_t \times D}$ and aerial image features $\mathbf{A} \in \mathbb{R}^{B \times N_a \times D}$, we introduce a shared learnable query $\mathbf{Q} \in \mathbb{R}^{K \times D}$ to interact with both modalities and map them into a unified semantic space:
\begin{equation}
\mathbf{Q}_a = CrossFormer(\mathbf{Q}, \mathbf{A}, \mathbf{A}), 
\end{equation}
\begin{equation}
\mathbf{Q}_t = CrossFormer(\mathbf{Q}, \mathbf{T}, \mathbf{T}),
\end{equation}
where $CrossFormer$ denotes the cross-modal interaction module, composed of a cross-attention layer followed by multiple self-attention and feed-forward layers, producing modality-aware query representations $\mathbf{Q}_t, \mathbf{Q}_a \in \mathbb{R}^{B \times K \times D}$.

To measure the reliability of each query token, we use the global class token of the corresponding modality as a semantic reference, and quantify the token’s reliability with a learnable Gaussian scale parameter. 
Taking the image modality as an example, we first predict the Gaussian scale $\sigma$ from the global class token $\mathbf{A}^C \in \mathbb{R}^{D}$ using an MLP:
\begin{equation}
\log \boldsymbol{\sigma} = \mathrm{MLP}(\mathbf{A^C}), \quad
\boldsymbol{\sigma} = \exp(\log \boldsymbol{\sigma}).
\end{equation}
Next, for each query token, we compute its overall similarity to the class token $r_j$, and map it to a membership degree $\mu^a_j$ using a Gaussian function. This membership degree reflects the reliability of the token in the image modality:
\begin{equation}
r_j = \frac{Q_a^{(j)} \cdot A^C}{|Q_a^{(j)}|_2 |A^C|_2}, \quad
\mu^a_j = \exp\Big(-\frac{(1 - r_j)^2}{2\sigma^2}\Big).
\end{equation}

The membership $\mu^t_j$ for the text modality is computed similarly. In this way, each token’s membership $\mu_j \in [0,1]$ not only reflects its alignment with the global semantic representation, but also quantifies the token’s reliability for cross-modal alignment, the higher the value, the more the token can be trusted; the lower the value, the more likely it is noisy or uninformative and should be suppressed.

We then fuse token-level memberships from both modalities using a fuzzy logic AND operation:
\begin{equation}
\mu^{\mathrm{joint}}_j = \mu^a_j \cdot \mu^t_j,
\end{equation}
where we adopt the multiplicative form as a differentiable soft AND operator, which emphasizes tokens with high confidence in both modalities while allowing gradient-based optimization.
Only tokens with high reliability in both modalities retain strong influence during cross-modal alignment. 
The token-level cosine similarity is then weighted by the joint membership to obtain the sample-level similarity:
\begin{equation}
s_j = \frac{{q_a^{(j)}}^\top q_t^{(j)}}{|q_v^{(j)}|_2 |q_t^{(j)}|_2}, \quad
\mathrm{sim}(Q_a,Q_t) = \frac{1}{K} \sum_{j=1}^{K} \mu^{\mathrm{joint}}_j s_j.
\end{equation}

The text-image similarity $\mathrm{sim}(Q_t,Q_a)$ is computed similarly. 
Using these bidirectional weighted similarities, we further employ the similarity distribution matching~\cite{jiang2023cross} to compute the fuzzy token alignment loss $L_{\mathrm{FTA}}$.
\begin{equation}
    p_{i,j} = \frac{\exp(\mathrm{sim}(Q_a^i, Q_t^j)/\tau)}{\sum_{k=1}^{N} \exp(\mathrm{sim}(Q_a^i, Q_t^k)/\tau)}, \quad
q_{i,j} = \frac{y_{i,j}}{\sum_{k=1}^{N} y_{i,k}},
\end{equation}
\begin{equation}
    \mathcal{L}_{\mathrm{FTA}} = \frac{1}{2} \sum_{i=1}^{N} \sum_{j=1}^{N}
\Bigg[ p_{i,j} \log \frac{p_{i,j}}{q_{i,j} + \epsilon} + p_{j,i} \log \frac{p_{j,i}}{q_{j,i} + \epsilon} \Bigg],
\end{equation}
where $\tau$ is a temperature hyperparameter, $\epsilon$ is used to prevent numerical instability during computation.
This mechanism ensures that only tokens that are reliably present in both modalities contribute significantly to alignment, suppressing noisy or weak tokens and enabling precise and robust fine-grained cross-modal alignment.

\section{Benchmark}
\begin{table*}[]
\caption{Comparison AERI-PEDES with other datasets for text-image person retrieval.}
\setlength{\tabcolsep}{0.2pt}{
\begin{tabular}{cc|c|c|c|c|c}
\hline
\multicolumn{2}{c|}{\multirow{2}{*}{Datasets}}                           & \multirow{2}{*}{AERI-PEDES} & \multirow{2}{*}{TBAPR~\cite{wang2025aea}} & \multirow{2}{*}{CUHK-PEDES~\cite{li2017person}} & \multirow{2}{*}{ICFG-PEDES~\cite{ding2021semantically}} & \multirow{2}{*}{RSTPReid~\cite{zhu2021dssl}} \\
\multicolumn{2}{c|}{}                                                    &                             &                        &                             &                             &                           \\ \hline
\multicolumn{1}{c|}{\multirow{3}{*}{Image}}   & Number  &  144,548  &  65,880 & 40,206 & 54,522  &  20,505   \\ 
\multicolumn{1}{c|}{}                         & Ground \& Aerial    &  $\surd$   &        $\surd$    &  $\times$  &    $\times$  &      $\times$ \\ 
\multicolumn{1}{c|}{}                         & Multi-Sources  & $\surd$ &  $\surd$  & $\surd$  & $\times$ & $\times$ \\ \hline
\multicolumn{1}{c|}{\multirow{3}{*}{Caption}} & Average Length  &  38.6  & 56.1  & 23.5  & 37.2 & 25.8  \\ 
\multicolumn{1}{c|}{}                         & Maximum Length &  105  &  87  &  96  &  83  &  70    \\
\multicolumn{1}{c|}{}                         & Generated (Train) \& Human (Test) &   $\surd$           &  $\times$  &  $\times$   &   $\times$    &   $\times$     \\  \hline
\end{tabular}
}
\label{tab_1}
\end{table*}

\begin{figure*}[!t]
\centering
\includegraphics[width=6.8in,height=1.6in]{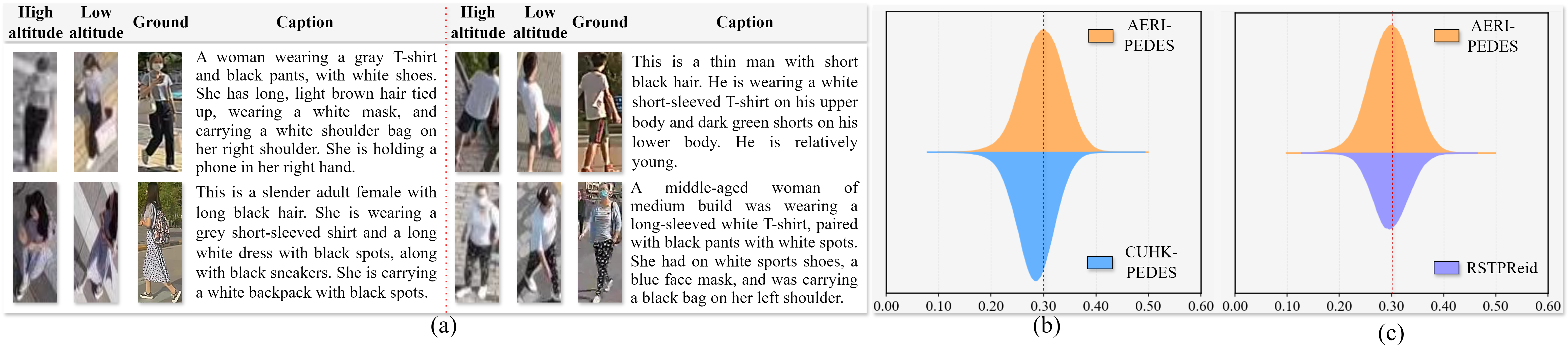}
\caption{(a) Illustrates sample data from the AERI-PEDES benchmark.
(b) Shows the text-image similarity distribution in comparison with CUHK-PEDES.
(c) Shows the text-image similarity distribution in comparison with RSTPReid.}
\label{fig_3}
\end{figure*}
\subsection{Overview}
To advance research on text-aerial person retrieval, we construct a large-scale benchmark, AERI-PEDES. 
The person images in this dataset are collected from three distinct aerial-ground person re-identification datasets~\cite{nguyen2023ag,nguyen2024ag,nguyen2025ag}, where each identity naturally contains multiple images from both ground and aerial viewpoints. 
To reduce manual annotation costs while ensuring the accuracy of captions, we propose a Chain-of-Though (CoT) based caption generation framework to produce high-quality captions.
In the test set, all captions are manually annotated to provide a reliable evaluation benchmark that closely reflects real-world scenarios.
Figure~\ref{fig_3}(a) illustrates several examples from AERI-PEDES.

\subsection{CoT-based Caption Genertaion Framework}
\begin{figure}[!t]
\centering
\includegraphics[width=3.2in,height=2.3in]{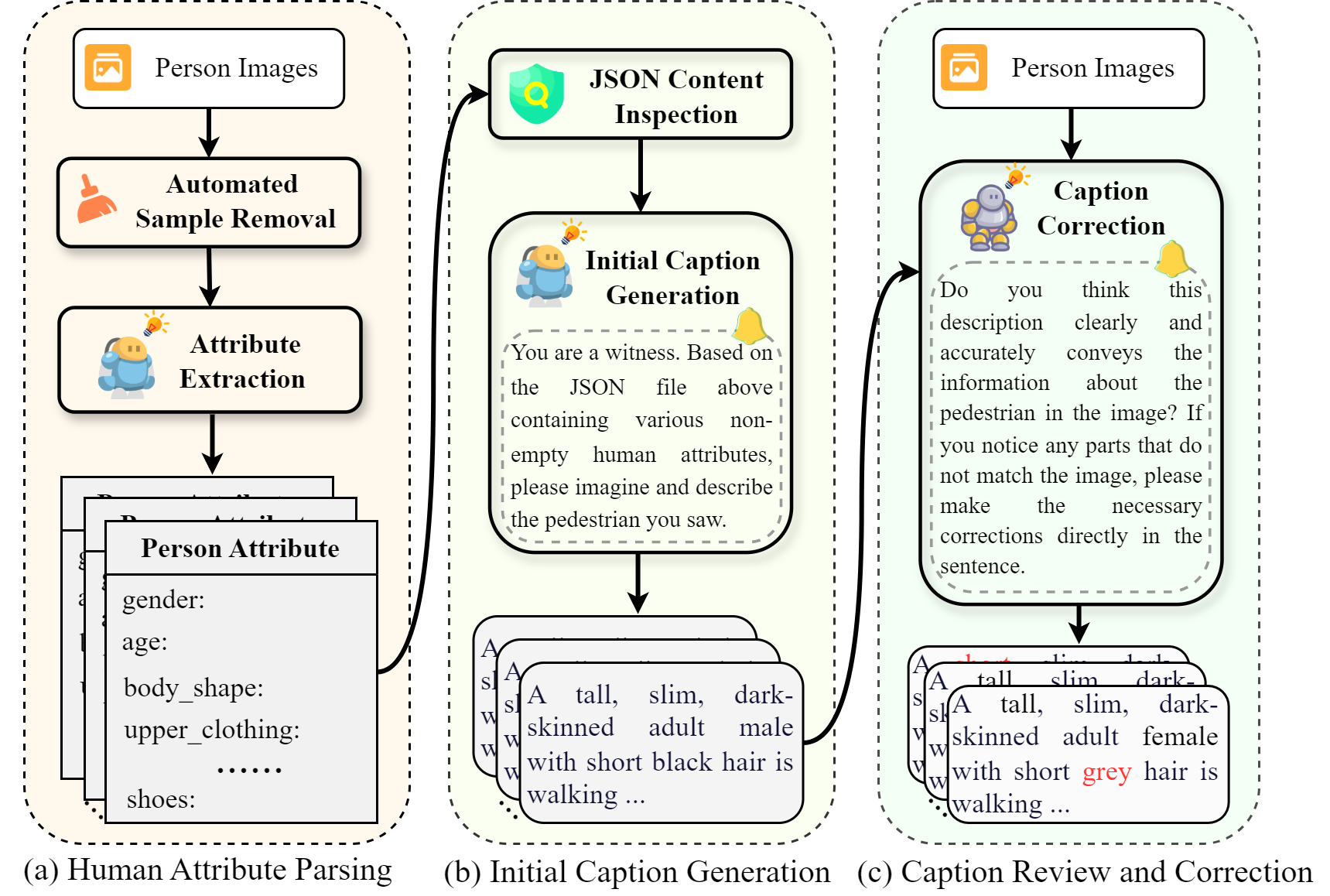}
\caption{Chain-of-Thought Based Caption Generation Framework.}
\label{fig_4}
\end{figure}
Existing caption generation methods based on multimodal large language models often suffer from attribute omission and visual hallucination. To address these issues, we propose a CoT based generation framework, as illustrated in Figure~\ref{fig_4}, which improves both the granularity and verifiability of generated captions.
Specifically, a perception model $f_{\text{percep}}$ first performs structured visual parsing on the input image $I$, extracting visible attributes $\mathbf{A} = {a_1, \dots, a_n}$ along with their supporting visual evidence $\mathbf{V}$ and confidence scores $(\mathbf{C}$, forming an intermediate representation $T_1 = (\mathbf{A}, \mathbf{V}, \mathbf{C})$. These attributes are then organized into an initial caption $S_{\text{init}}$ with a recorded reasoning trace, yielding $T_2 = (S_{\text{init}}, \text{trace}_{\text{gen}})$.
To correct potential omissions or errors, a secondary model $f_{\text{corr}}$ performs vision-guided auditing on $(I, S_{\text{init}})$, producing a refined caption $S_{\text{refined}}$ and a correction trace, resulting in the final state $T_3 = (S_{\text{refined}}, \text{trace}_{\text{corr}})$.
This progressive CoT mechanism effectively integrates visual parsing and auditing to preserve fine-grained attribute details while ensuring that the generated captions remain consistent with visual evidence. Further implementation details are provided in the \textbf{supplementary material}.

\subsection{Benchmark Properties and Statistics}
We construct AERI-PEDES, a large-scale cross-view benchmark with 4,659 identities, generated using a CoT-based captioning framework.
The training set includes 3,659 identities, 112,672 aerial images, 26,351 ground images, and 26,213 generated captions, while the test set contains 1,000 identities, 5,525 aerial images, and 6,141 manually annotated captions for reliable evaluation.
As shown in Table~\ref{tab_1}, compared with other person retrieval datasets, AERI-PEDES offers larger scale, broader cross-platform coverage, and richer scene diversity. The captions are detailed, with a maximum length of 108 words and an average of 38.6 words; only the training captions are generated, whereas the test captions remain manually annotated to better reflect real-world usage.
We further analyze the CLIP-based image–text similarity distributions (Figure~\ref{fig_3}(b)(c)).
Unlike fully manual datasets such as CUHK-PEDES and RSTPReid, AERI-PEDES exhibits a wider similarity spread, demonstrating the effectiveness of the CoT captioning framework in producing high-quality, semantically consistent descriptions.

\section{Experiments}
\begin{table*}[htbp]
\setlength{\tabcolsep}{2.6pt}
\caption{Comparison results on the AERI-PEDES, and TBAPR datasets. ~\label{tab:table2}}
\centering
\begin{tabular}{cc|ccccc|ccccc}
\hline
\multirow{2}{*}{Method} & \multirow{2}{*}{Ref}  & \multicolumn{5}{c|}{AERI-PEDES} & \multicolumn{5}{c}{TBAPR} \\ 
\cline{3-12} 
& & Rank-1 & Rank-5 & Rank-10 & mAP & RSum & Rank-1 & Rank-5 & Rank-10 & mAP & RSum \\
\hline
IRRA~\cite{jiang2023cross} & CVPR23 & 35.14 & 53.23  & 63.19  &  33.42& 151.57 & 39.63 & 58.72 & 67.69 & 35.31 & 166.04 \\
APTM~\cite{yang2023towards}&  MM23 & 34.62  & 53.95   & 64.5      & 31.09 & 153.07& 43.59 & 62.03 & 69.75 & 38.71 & 175.37   \\
RDE~\cite{qin2024noisy}& CVPR24   & 38.56  & 58.26   & 67.89     & 37.16& 164.71& 37.31 & 54.06 & 60.75 & 32.17  &  152.12 \\
CFAM~\cite{zuo2024ufinebench} & CVPR24 & 30.77  & 51.37   & 61.61    & 30.4 & 143.75 & 48.34& 66.31&73.21& 42.67 & 184.78    \\
NAM~\cite{tan2024harnessing} & CVPR24 & 42.47 & 61.72  & 69.99   & 40.22 & 174.17&  46.56 & 63.13 & 70.13 & 40.92 & 179.82  \\
VFE~\cite{shen2025enhancing} & KBS25 & 35.76  & 55.35  & 65.56   & 35.35& 156.67& 47.94 & 62.5 & 68.17 & 42.18 & 178.63 \\
DM-Adpeter~\cite{liu2025dm}& AAAI25 &  33.42 & 53.17  & 62.79   & 32.41 & 149.37& 37.81& 58.34&66.56& 33.28 & 162.71 \\
LPNC~\cite{wang2025learnable}& TIFS25 & 35.65  & 53.61   & 63.69    & 35.19 & 152.95& 41.78 & 58.03 & 65.50 & 37.87 & 165.31 \\
LPNC+Pretrain~\cite{wang2025learnable}& TIFS25 & 43.79  & 61.49   & 70.40      & 42.22& 175.68 & 45.41 & 62.31 & 69.94 & 42.17 & 177.66 \\
AEA-FIRM~\cite{wang2025aea}&  TCSVT25 &37.94&56.66&65.71& 34.89 & 160.31 & 44.75 & 62.38 & 69.13 & 36.28& 176.26\\
%Pretrain on HAM
AEA-FIRM+Pretrain~\cite{jiang2025modeling}&  TCSVT25 &44.42& 61.96& 71.03& 41.12 & 177.41 & 48.15& 63.87& 71.21&42.01& 183.23\\
HAM~\cite{jiang2025modeling} & CVPR25 & 44.58  & 63.52   & 72.67     & 42.45  & 180.77 & 47.81 & 64.96 & 72.53 & 41.86 & 185.30 \\
\hline
Ours (W/O Ground) & - & 45.06 & 64.53  & 73.21 & 43.27 & 182.80 & 49.15  & 65.88 & 73.47 & 42.89 & 188.50  \\
Ours (With Ground) & - & 47.16 & 65.66 & 73.83 & 44.79 & 186.65 & 49.47 & 66.50  & 73.06  & 43.96 & 189.03   \\
\hline
\end{tabular}
\end{table*}
\subsection{Datasets and Evaluation Metrics}
\textbf{Datasets.} We evaluate our method on the AERI-PEDES and TBAPR~\cite{wang2025aea}. 

\noindent\textbf{TBAPR} is the first benchmark specifically designed for text-aerial person retrieval. It contains 65,880 person images, with the training set comprising 1,180 identities and the test set containing 529 identities. 

\noindent\textbf{Evaluation Metrics.} We evaluate our method using multiple widely adopted retrieval metrics, including Rank-k (k=1, 5, 10), mAP (Mean Average Precision) and RSum (Sum of Rank),  detailed introductions are provided in the \textbf{supplementary material}.

\subsection{Implementation Details}
CFAN uses the CLIP text encoder for textual feature extraction, while aerial and ground images share a CLIP-based visual encoder initialized with the pre-trained weights provided by Jiang et al~\cite{jiang2025modeling}. 
Image data are processed with standard augmentation operations and uniformly resized to $384 \times 128$, while text inputs are enhanced via random masking to improve model robustness.
The cross-modal interaction query in the FTA module consists of 4 learnable 512-dimensional tokens. 
The model is trained using the Adam optimizer for 60 epochs with an initial learning rate of $5 \times 10^{-6}$ and a cosine decay schedule.
The batch size is set to 64, and training is conducted on a single RTX 4090 GPU.

\subsection{Comparison with State-of-the-Art Methods}
\noindent\textbf{Performance Comparison on the AERI-PEDES Datasets.}
As shown in Table~\ref{tab:table2}, we conduct a comprehensive evaluation of the proposed method on the AERI-PEDES dataset and perform detailed comparisons with existing state-of-the-art approaches. Specifically, AEA-FIRM, as the first attempt targeting text–aerial person alignment, demonstrates certain advantages over traditional text–image person retrieval methods on AERI-PEDES. However, due to the extremely large viewpoint variations in this benchmark, the difficulty of cross-modal alignment is significantly amplified. Even with strong pretrained models, AEA-FIRM only achieves 44.42\% Rank-1 accuracy, 41.12\% mAP, and 177.41\% RSum, performing on par with HAM.
In contrast, our method delivers substantial performance improvements. Under the non-ground-assisted setting, it already surpasses all competing approaches. When ground-view images are further incorporated as auxiliary cues, our method sets a new state of the art, achieving 47.16\% Rank-1, 44.79\% mAP, and 186.65\% RSum, representing nearly a 6\% gain in RSum over the previous best method.
These results highlight the superior capability of our approach in handling cross-modal alignment under extreme viewpoint disparities. While existing methods struggle in this challenging scenario, our model benefits from fuzzy logic based token-level reliability modeling, enabling more accurate and robust fine-grained cross-modal alignment.

\noindent\textbf{Performance Comparison on the TBAPR Datasets.}
On the TBAPR dataset, our method consistently outperforms all compared approaches across all evaluation metrics. Specifically, without using ground-view images, our method already achieves 49.15\% Rank-1 accuracy and 42.89\% mAP, surpassing all competing methods. When ground-view images are further incorporated as auxiliary information, our method reaches a new SoTA, achieving 66.50\% Rank-5 and 189.03\% RSum.
It is worth noting that, since the TBAPR dataset contains many low-altitude aerial images, the visual differences between aerial and ground-view images are relatively small. This allows direct alignment between text and aerial images to achieve strong performance, limiting the auxiliary contribution of ground images. In contrast, our method can further improve results thanks to the CDA module, which adaptively balances the contributions of text–aerial and text–ground alignment. This enables robust cross-modal semantic matching and effectively leveraging ground-view information.

\subsection{Ablation Study}
In this section, we evaluate each component of CFAN using the baseline model with ground-image assisted bridge alignment loss for comparison.
\begin{table}[t]
\renewcommand{\arraystretch}{1}
\setlength{\tabcolsep}{0.1cm}{
\centering
\caption{Ablation experiments of different components on AERI-PEDES dataset.~\label{tab:table3}}
\begin{tabular}{l|ccccc}
\hline
Methods & Rank-1 & Rank-5 & Rank-10 & mAP & RSum \\
\hline
Baseline & 43.88 & 61.11 & 69.85 & 41.58 & 174.84  \\
+ CDA & 46.18  & 64.40 & 72.46 & 43.98 & 183.04 \\
+ FTA & 44.55 & 61.77 & 70.32 & 41.89 & 176.64 \\
+ CDA + FTA & 47.16 & 65.66 & 73.83 & 44.79 & 186.65\\
\hline
\end{tabular}
}
\end{table}

\begin{table}[t]
\renewcommand{\arraystretch}{1}
\setlength{\tabcolsep}{0.18cm}{
\centering
\caption{Impact of Different Bridge Modalities in CDA.~\label{tab:table4}}
\begin{tabular}{c|ccccc}
\hline
Bridge & Rank-1 & Rank-5 & Rank-10 & mAP & RSum \\
\hline
None & 45.06 & 64.53 & 73.21 & 43.27 & 182.80  \\
Aerial & 46.08  & 64.60 & 73.33 & 44.20 & 184.01 \\
Ground & 47.16 & 65.66 & 73.83 & 44.79 & 186.65 \\
\hline
\end{tabular}
}
\end{table}

\noindent\textbf{Effectiveness of the Context-Aware Dynamic Alignment Loss.}
As shown in Table~\ref{tab:table3}, using only bridge alignment (baseline) achieves 43.88\% Rank-1, 41.58\% mAP, and 174.84\% RSum. However, although bridge alignment alleviates the semantic gap between aerial images and text, it fails to fully exploit direct alignment for low-difficulty samples, leading to suboptimal matching performance.
After introducing the CDA loss, key metrics improve significantly, with RSum increasing by 8.2\%, along with corresponding gains in Rank-1 and mAP.
This improvement is mainly attributed to CDA, which dynamically estimates alignment difficulty and adaptively balances direct and bridge alignment. Specifically, low-difficulty samples emphasize direct alignment, while high-difficulty samples rely more on bridge alignment, thereby enhancing overall alignment accuracy and robustness.

\noindent\textbf{Effectiveness of the Fuzzy Token Alignment.}
As shown in Table~\ref{tab:table3}, incorporating the FTA module into the baseline yields improvements of 0.67\%, 0.31\%, and 1.2\% in Rank-1, mAP, and RSum, respectively.
Further integrating FTA on top of the CDA loss leads to additional gains of 0.98\%, 0.81\%, and 3.61\%.
These improvements mainly stem from the Fuzzy Token Alignment module, which modulates fine-grained token correspondences through fuzzy membership and suppresses interference from unobservable or noisy tokens during token-level alignment.
Consequently, only reliable and semantically supported tokens participate in cross-modal matching, resulting in more accurate alignment.

\noindent\textbf{Impact of Different Bridge in CDA.}
As shown in Table~\ref{tab:table4}, “None” indicates no intermediate bridge, “Aerial” uses low-altitude aerial images of the same identity, and “Ground” uses ground-level images. Introducing a bridge (Aerial or Ground) consistently improves performance, showing that intermediate modalities help ease cross-modal alignment. Aerial images provide noticeable gains, while Ground images achieve the best results due to higher semantic consistency with text. The small gap between Aerial and Ground suggests that CDA does not strictly rely on ground images and can flexibly use different bridge modalities for robust alignment.

\subsection{Parameter Sensitivity Analysis}
\begin{figure}[!t]
\centering
\includegraphics[width=3.2in,height=1.7in]{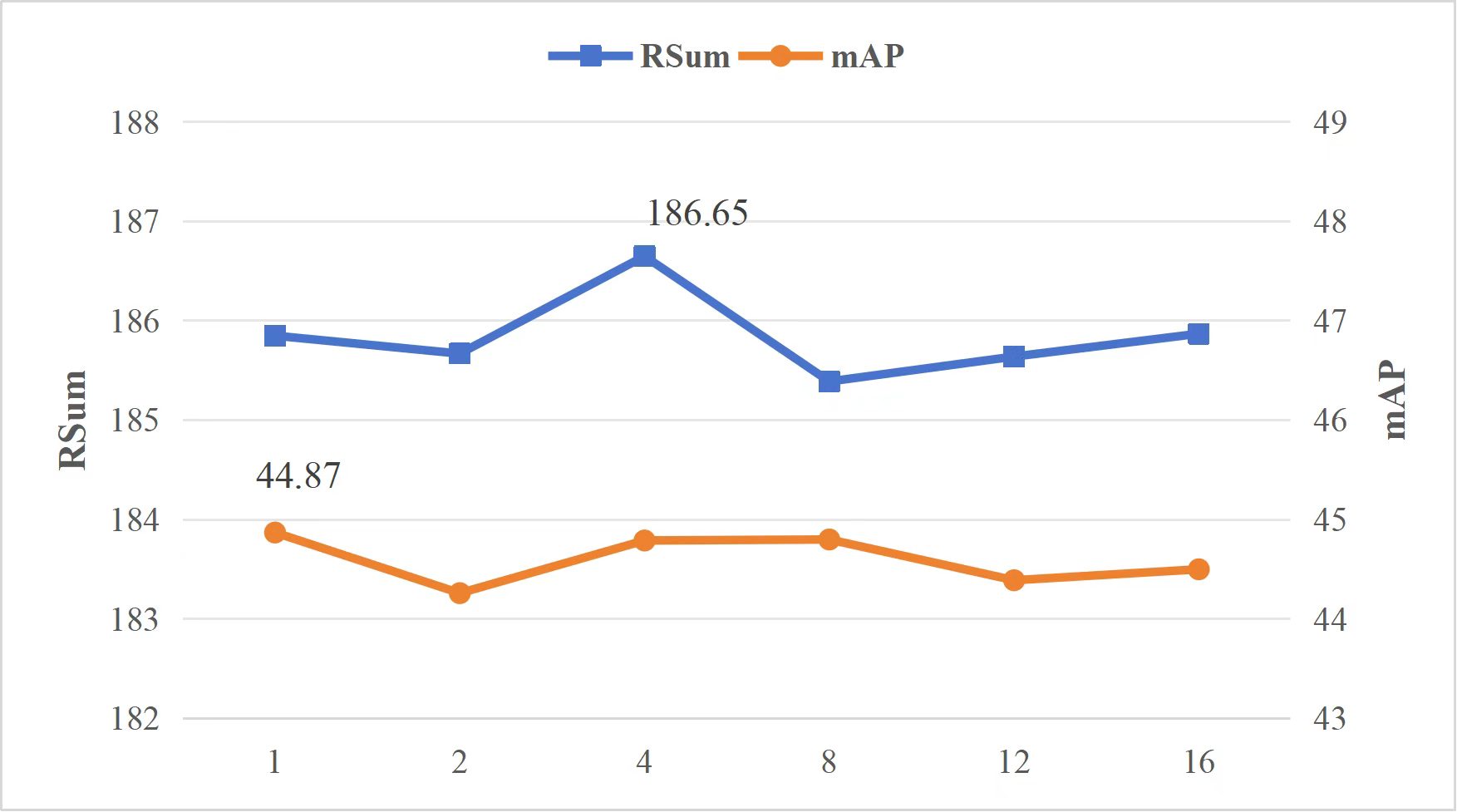}
\caption{Impact of the Number of Learnable Query Tokens.}
\label{fig_6}
\end{figure}
\begin{table}[t]
\renewcommand{\arraystretch}{1}
\setlength{\tabcolsep}{0.09cm}{
\centering
\caption{Sensitivity analysis of the hyperparameter $k$ in the context-aware dynamic alignment module.~\label{tab:table5}}
\begin{tabular}{c|ccccccc}
\hline
$k$ & 1 & 2 & 4 & 8 & 12 & 16 \\
\hline
Rank-1 & 47.16  & 46.91  & 47.04 & 46.77  & 46.23 & 46.04 & \\
mAP & 44.79 & 44.58  & 44.50 & 43.83  & 43.70 & 43.55 & \\
RSum & 186.65 & 185.88 & 185.91 & 184.11& 183.99 & 183.31 & \\
\hline
\end{tabular}
}
\end{table}
We conduct an ablation study on the number of tokens in the learnable query $Q$. As shown in Figure~\ref{fig_6}, we evaluated the effect of varying the token count from 1 to 16. The results indicate that changes in token number have little impact on the mAP metric. When the number of token is set to 4, the model achieves the highest RSum performance of 186.65\%.
However, further increasing the number of tokens leads to performance degradation, likely due to redundancy and interference among tokens caused by over-parameterization, which weakens fine-grained cross-modal alignment. Based on these observations, we set the number of learnable query tokens to 4 in our final configuration.

We further analyze the sensitivity of the hyperparameter $k$ in the CDA module. As shown in Table~\ref{tab:table5}, the best performance is achieved at $k$, while larger $k$ values lead to a gradual decline. This is because increasing $k$ makes the Sigmoid function overly steep, resulting in overly extreme weighting between direct and bridge alignment, which weakens fine-grained adjustment and degrades overall performance.

%% file: sec/3_finalcopy.tex
\section{Conclusion}
In this paper, we propose a Cross-modal Fuzzy Alignment Network for text-aerial person retrieval, aimed at addressing the challenges of cross-modal alignment arising from degraded visual cues in aerial images.
Firstly, we introduce a Fuzzy Token Alignment module, which leverages fuzzy logic to quantify token-level reliability, enabling accurate and robust fine-grained alignment between text and aerial image features.
Secondly, we design a Context-Aware Dynamic Alignment module, which incorporates ground-view images as a bridging mechanism to effectively reduce alignment discrepancies between text and aerial images, adaptively balancing direct and bridge-assisted alignment for each sample.
In addition, we construct AERI-PEDES, a large-scale benchmark dataset generated with a chain-of-thought approach. Multi-stage prompts produce attribute parsing, initial captions, and refined descriptions with traceable intermediate states, ensuring high accuracy, completeness, and visual consistency of the generated captions.
Extensive experiments demonstrate the superior effectiveness and robustness of our method.

%% file: main.bib
@String(AAAI = {AAAI})

@article{yan2023clip,
  title={Clip-driven fine-grained text-image person re-identification},
  author={Yan, Shuanglin and Dong, Neng and Zhang, Liyan and Tang, Jinhui},
  journal={IEEE Transactions on Image Processing},
  volume={32},
  pages={6032--6046},
  year={2023},
  publisher={IEEE}
}

@inproceedings{jiang2023cross,
  title={Cross-modal implicit relation reasoning and aligning for text-to-image person retrieval},
  author={Jiang, Ding and Ye, Mang},
  booktitle={Proceedings of the IEEE/CVF Conference on Computer Vision and Pattern Recognition},
  pages={2787--2797},
  year={2023}
}

@inproceedings{radford2021learning,
  title={Learning transferable visual models from natural language supervision},
  author={Radford, Alec and Kim, Jong Wook and Hallacy, Chris and Ramesh, Aditya and Goh, Gabriel and Agarwal, Sandhini and Sastry, Girish and Askell, Amanda and Mishkin, Pamela and Clark, Jack and others},
  booktitle={International conference on machine learning},
  pages={8748--8763},
  year={2021},
  organization={PmLR}
}

@inproceedings{tan2024harnessing,
  title={Harnessing the power of mllms for transferable text-to-image person reid},
  author={Tan, Wentan and Ding, Changxing and Jiang, Jiayu and Wang, Fei and Zhan, Yibing and Tao, Dapeng},
  booktitle={Proceedings of the IEEE/CVF Conference on Computer Vision and Pattern Recognition},
  pages={17127--17137},
  year={2024}
}

@article{li2021align,
  title={Align before fuse: Vision and language representation learning with momentum distillation},
  author={Li, Junnan and Selvaraju, Ramprasaath and Gotmare, Akhilesh and Joty, Shafiq and Xiong, Caiming and Hoi, Steven Chu Hong},
  journal={Advances in neural information processing systems},
  volume={34},
  pages={9694--9705},
  year={2021}
}

@inproceedings{li2017person,
  title={Person search with natural language description},
  author={Li, Shuang and Xiao, Tong and Li, Hongsheng and Zhou, Bolei and Yue, Dayu and Wang, Xiaogang},
  booktitle={Proceedings of the IEEE conference on computer vision and pattern recognition},
  pages={1970--1979},
  year={2017}
}

@inproceedings{zhu2021dssl,
  title={Dssl: Deep surroundings-person separation learning for text-based person retrieval},
  author={Zhu, Aichun and Wang, Zijie and Li, Yifeng and Wan, Xili and Jin, Jing and Wang, Tian and Hu, Fangqiang and Hua, Gang},
  booktitle={Proceedings of the 29th ACM international conference on multimedia},
  pages={209--217},
  year={2021}
}

@article{ding2021semantically,
  title={Semantically self-aligned network for text-to-image part-aware person re-identification},
  author={Ding, Zefeng and Ding, Changxing and Shao, Zhiyin and Tao, Dacheng},
  journal={arXiv preprint arXiv:2107.12666},
  year={2021}
}

@inproceedings{wang2020vitaa,
  title={Vitaa: Visual-textual attributes alignment in person search by natural language},
  author={Wang, Zhe and Fang, Zhiyuan and Wang, Jun and Yang, Yezhou},
  booktitle={Computer vision--ECCV 2020: 16th European conference, glasgow, UK, August 23--28, 2020, proceedings, part XII 16},
  pages={402--420},
  year={2020},
  organization={Springer}
}

@article{chen2022tipcb,
  title={TIPCB: A simple but effective part-based convolutional baseline for text-based person search},
  author={Chen, Yuhao and Zhang, Guoqing and Lu, Yujiang and Wang, Zhenxing and Zheng, Yuhui},
  journal={Neurocomputing},
  volume={494},
  pages={171--181},
  year={2022},
  publisher={Elsevier}
}

@article{nguyen2024ag,
  title={AG-ReID. v2: Bridging aerial and ground views for person re-identification},
  author={Nguyen, Huy and Nguyen, Kien and Sridharan, Sridha and Fookes, Clinton},
  journal={IEEE Transactions on Information Forensics and Security},
  volume={19},
  pages={2896--2908},
  year={2024},
  publisher={IEEE}
}

@inproceedings{nguyen2025ag,
  title={AG-VPReID: A Challenging Large-Scale Benchmark for Aerial-Ground Video-based Person Re-Identification},
  author={Nguyen, Huy and Nguyen, Kien and Pemasiri, Akila and Liu, Feng and Sridharan, Sridha and Fookes, Clinton},
  booktitle={Proceedings of the Computer Vision and Pattern Recognition Conference},
  pages={1241--1251},
  year={2025}
}

@inproceedings{nguyen2023ag,
  title={Ag-reid 2023: Aerial-ground person re-identification challenge results},
  author={Nguyen, Kien and Fookes, Clinton and Sridharan, Sridha and Liu, Feng and Liu, Xiaoming and Ross, Arun and Michalski, Dana and Nguyen, Huy and Deb, Debayan and Kothari, Mahak and others},
  booktitle={2023 IEEE International Joint Conference on Biometrics (IJCB)},
  pages={1--10},
  year={2023},
  organization={IEEE}
}

@article{wang2025aea,
  title={AEA-FIRM: Adaptive Elastic Alignment with Fine-Grained Representation Mining for Text-based Aerial Pedestrian Retrieval},
  author={Wang, Yihao and Yang, Meng and Cao, Rui and Gao, Guangwei},
  journal={IEEE Transactions on Circuits and Systems for Video Technology},
  year={2025},
  publisher={IEEE}
}

@inproceedings{yang2023towards,
  title={Towards unified text-based person retrieval: A large-scale multi-attribute and language search benchmark},
  author={Yang, Shuyu and Zhou, Yinan and Zheng, Zhedong and Wang, Yaxiong and Zhu, Li and Wu, Yujiao},
  booktitle={Proceedings of the 31st ACM International Conference on Multimedia},
  pages={4492--4501},
  year={2023}
}

@inproceedings{zuo2024ufinebench,
  title={Ufinebench: Towards text-based person retrieval with ultra-fine granularity},
  author={Zuo, Jialong and Zhou, Hanyu and Nie, Ying and Zhang, Feng and Guo, Tianyu and Sang, Nong and Wang, Yunhe and Gao, Changxin},
  booktitle={Proceedings of the IEEE/CVF Conference on Computer Vision and Pattern Recognition},
  pages={22010--22019},
  year={2024}
}

@inproceedings{qin2024noisy,
  title={Noisy-correspondence learning for text-to-image person re-identification},
  author={Qin, Yang and Chen, Yingke and Peng, Dezhong and Peng, Xi and Zhou, Joey Tianyi and Hu, Peng},
  booktitle={Proceedings of the IEEE/CVF Conference on Computer Vision and Pattern Recognition},
  pages={27197--27206},
  year={2024}
}

@inproceedings{liu2025dm,
  title={Dm-adapter: Domain-aware mixture-of-adapters for text-based person retrieval},
  author={Liu, Yating and Liu, Zimo and Lan, Xiangyuan and Yang, Wenming and Li, Yaowei and Liao, Qingmin},
  booktitle={Proceedings of the AAAI Conference on Artificial Intelligence},
  volume={39},
  number={6},
  pages={5703--5711},
  year={2025}
}

@article{shen2025enhancing,
  title={Enhancing visual representation for text-based person searching},
  author={Shen, Wei and Fang, Ming and Wang, Yuxia and Xiao, Jiafeng and Li, Diping and Chen, Huangqun and Xu, Ling and Zhang, Weifeng},
  journal={Knowledge-Based Systems},
  volume={309},
  pages={112893},
  year={2025},
  publisher={Elsevier}
}

@article{wang2025learnable,
  title={Learnable Prompts With Neighbor-Aware Correction for Text-Based Person Search},
  author={Wang, Xueping and Wu, Hao and Liu, Min and Wang, Yaonan},
  journal={IEEE Transactions on Information Forensics and Security},
  year={2025},
  publisher={IEEE}
}

@inproceedings{jiang2025modeling,
  title={Modeling Thousands of Human Annotators for Generalizable Text-to-Image Person Re-identification},
  author={Jiang, Jiayu and Ding, Changxing and Tan, Wentao and Wang, Junhong and Tao, Jin and Xu, Xiangmin},
  booktitle={Proceedings of the Computer Vision and Pattern Recognition Conference},
  pages={9220--9230},
  year={2025}
}

@inproceedings{jing2020pose,
  title={Pose-guided multi-granularity attention network for text-based person search},
  author={Jing, Ya and Si, Chenyang and Wang, Junbo and Wang, Wei and Wang, Liang and Tan, Tieniu},
  booktitle={Proceedings of the AAAI Conference on Artificial Intelligence},
  volume={34},
  number={07},
  pages={11189--11196},
  year={2020}
}

@inproceedings{deng2025learning,
title={Learning Hierarchical Cross-modal Association with Intra-modal Context for Text-Image Person Retrieval},
author={Deng, Yifei and Li, Chenglong and Wang, Futian and Tang, Jin},
booktitle={Proceedings of the 33rd ACM International Conference on Multimedia},
pages={2723--2731},
year={2025}
}

@article{deng2025uncertainty,
  title={Uncertainty-aware coarse-to-fine alignment for text-image person retrieval},
  author={Deng, Yifei and Chen, Zhengyu and Li, Chenglong and Tang, Jin},
  journal={Visual Intelligence},
  volume={3},
  number={1},
  pages={6},
  year={2025},
  publisher={Springer}
}

@article{eom2019learning,
  title={Learning disentangled representation for robust person re-identification},
  author={Eom, Chanho and Ham, Bumsub},
  journal={Advances in neural information processing systems},
  volume={32},
  year={2019}
}

@article{wei2022chain,
  title={Chain-of-thought prompting elicits reasoning in large language models},
  author={Wei, Jason and Wang, Xuezhi and Schuurmans, Dale and Bosma, Maarten and Xia, Fei and Chi, Ed and Le, Quoc V and Zhou, Denny and others},
  journal={Advances in neural information processing systems},
  volume={35},
  pages={24824--24837},
  year={2022}
}

@article{zheng2020dual,
  title={Dual-path convolutional image-text embeddings with instance loss},
  author={Zheng, Zhedong and Zheng, Liang and Garrett, Michael and Yang, Yi and Xu, Mingliang and Shen, Yi-Dong},
  journal={ACM Transactions on Multimedia Computing, Communications, and Applications (TOMM)},
  volume={16},
  number={2},
  pages={1--23},
  year={2020},
  publisher={ACM New York, NY, USA}
}

@inproceedings{li2023blip,
  title={Blip-2: Bootstrapping language-image pre-training with frozen image encoders and large language models},
  author={Li, Junnan and Li, Dongxu and Savarese, Silvio and Hoi, Steven},
  booktitle={International conference on machine learning},
  pages={19730--19742},
  year={2023},
  organization={PMLR}
}

@article{zadeh1965fuzzy,
  title={Fuzzy sets},
  author={Zadeh, Lotfi A},
  journal={Information and control},
  volume={8},
  number={3},
  pages={338--353},
  year={1965},
  publisher={Elsevier}
}

@article{ieracitano2022fuzzy,
  title={A fuzzy-enhanced deep learning approach for early detection of Covid-19 pneumonia from portable chest X-ray images},
  author={Ieracitano, Cosimo and Mammone, Nadia and Versaci, Mario and Varone, Giuseppe and Ali, Abder-Rahman and Armentano, Antonio and Calabrese, Grazia and Ferrarelli, Anna and Turano, Lorena and Tebala, Carmela and others},
  journal={Neurocomputing},
  volume={481},
  pages={202--215},
  year={2022},
  publisher={Elsevier}
}

@article{golcuk2022hybrid,
  title={Hybrid fuzzy expert system and difference equation software filter for biomedical sensors},
  author={Golcuk, Adem},
  journal={IEEE Transactions on Instrumentation and Measurement},
  volume={71},
  pages={1--12},
  year={2022},
  publisher={IEEE}
}

@inproceedings{duan2025fuzzy,
  title={Fuzzy multimodal learning for trusted cross-modal retrieval},
  author={Duan, Siyuan and Sun, Yuan and Peng, Dezhong and Liu, Zheng and Song, Xiaomin and Hu, Peng},
  booktitle={Proceedings of the Computer Vision and Pattern Recognition Conference},
  pages={20747--20756},
  year={2025}
}

@article{ding2021multimodal,
  title={Multimodal infant brain segmentation by fuzzy-informed deep learning},
  author={Ding, Weiping and Abdel-Basset, Mohamed and Hawash, Hossam and Pedrycz, Witold},
  journal={IEEE Transactions on Fuzzy Systems},
  volume={30},
  number={4},
  pages={1088--1101},
  year={2021},
  publisher={IEEE}
}

@article{ahmadi2021fwnnet,
  title={FWNNet: Presentation of a New Classifier of Brain Tumor Diagnosis Based on Fuzzy Logic and the Wavelet-Based Neural Network Using Machine-Learning Methods},
  author={Ahmadi, Mohsen and Dashti Ahangar, Fatemeh and Astaraki, Nikoo and Abbasi, Mohammad and Babaei, Behzad},
  journal={Computational Intelligence and Neuroscience},
  volume={2021},
  number={1},
  pages={8542637},
  year={2021},
  publisher={Wiley Online Library}
}

@article{nan2023fuzzy,
  title={Fuzzy attention neural network to tackle discontinuity in airway segmentation},
  author={Nan, Yang and Del Ser, Javier and Tang, Zeyu and Tang, Peng and Xing, Xiaodan and Fang, Yingying and Herrera, Francisco and Pedrycz, Witold and Walsh, Simon and Yang, Guang},
  journal={IEEE transactions on neural networks and learning systems},
  volume={35},
  number={6},
  pages={7391--7404},
  year={2023},
  publisher={IEEE}
}

@article{xu2025residual,
  title={Residual Fuzzy Alignment on Hypergraph for Open-Set 3D Cross-Modal Retrieval},
  author={Xu, Yang and Feng, Yifan and Zhuang, Xu and Wang, Jason and Wu, Zongze and Gao, Yue},
  journal={IEEE Transactions on Multimedia},
  year={2025},
  publisher={IEEE}
}

@InProceedings{ZhaXia_Learning_MICCAI2025,
        author = { Zhao, Xiaowei AND Li, Chenglong AND Tang, Jin AND Li, Chuanfu},
        title = { { Learning with Explicit Topological Priors for Chest X-ray Rib Segmentation } },
        booktitle = {proceedings of Medical Image Computing and Computer Assisted Intervention -- MICCAI 2025},
        year = {2025},
        publisher = {Springer Nature Switzerland},
        volume = {LNCS 15975},
        month = {September},
        page = {300 -- 309}
}

@article{li2025ckdf,
  title={CKDF-V2: Effectively Alleviating Representation Shift for Continual Learning With Small Memory},
  author={Li, Kunchi and Chen, Hongyang and Wan, Jun and Yu, Shan},
  journal={IEEE Transactions on Neural Networks and Learning Systems},
  year={2025},
  publisher={IEEE}
}

@article{li2025enhance,
  title={Enhance the old representations’ adaptability dynamically for exemplar-free continual learning},
  author={Li, Kunchi and Ding, Chaoyue and Wan, Jun and Yu, Shan},
  journal={Neurocomputing},
  pages={130286},
  year={2025},
  publisher={Elsevier}
}

@article{pan2025diverse,
  title={Diverse feature generation for zero-shot chinese character recognition},
  author={Pan, Song-Liang and Li, Kunchi and Wang, Da-Han and Zhang, Xu-Yao and Liu, Jiantao and Zhu, Shunzhi},
  journal={Expert Systems with Applications},
  pages={129442},
  year={2025},
  publisher={Elsevier}
}

@article{jin2025sequencepar,
  title={Sequencepar: Understanding pedestrian attributes via a sequence generation paradigm},
  author={Jin, Jiandong and Wang, Xiao and Lin, Yin and Li, Chenglong and Huang, Lili and Zheng, Aihua and Tang, Jin},
  journal={Pattern Recognition},
  pages={112356},
  year={2025},
  publisher={Elsevier}
}

@inproceedings{jin2025pedestrian,
  title={Pedestrian attribute recognition: A new benchmark dataset and a large language model augmented framework},
  author={Jin, Jiandong and Wang, Xiao and Zhu, Qian and Wang, Haiyang and Li, Chenglong},
  booktitle={Proceedings of the AAAI Conference on Artificial Intelligence},
  volume={39},
  number={4},
  pages={4138--4146},
  year={2025}
}

@article{shen2025fine,
  title={Fine-grained preference optimization improves spatial reasoning in vlms},
  author={Shen, Yifan and Liu, Yuanzhe and Zhu, Jingyuan and Cao, Xu and Zhang, Xiaofeng and He, Yixiao and Ye, Wenming and Rehg, James Matthew and Lourentzou, Ismini},
  journal={arXiv preprint arXiv:2506.21656},
  year={2025}
}

@article{li2026toward,
  title={Toward Cognitive Supersensing in Multimodal Large Language Model},
  author={Li, Boyi and Shen, Yifan and Liu, Yuanzhe and Xu, Yifan and Liu, Jiateng and Li, Xinzhuo and Li, Zhengyuan and Zhu, Jingyuan and Zhong, Yunhan and Lan, Fangzhou and others},
  journal={arXiv preprint arXiv:2602.01541},
  year={2026}
}

@article{xu2025stare,
  title={STARE-VLA: Progressive Stage-Aware Reinforcement for Fine-Tuning Vision-Language-Action Models},
  author={Xu, Feng and Zhai, Guangyao and Kong, Xin and Fu, Tingzhong and Gordon, Daniel FN and An, Xueli and Busam, Benjamin},
  journal={arXiv preprint arXiv:2512.05107},
  year={2025}
}

@article{ding2025quality,
  title={Quality-Aware Spatio-Temporal Transformer Network for RGBT Tracking},
  author={Ding, Zhaodong and Li, Chenglong and Wang, Tao and Wang, Futian},
  journal={IEEE Transactions on Image Processing},
  volume={34},
  pages={7845--7858},
  year={2025},
  publisher={IEEE}
}

@Article {MIR-2022-05-167,
author = {Haoyu Lu and Yuqi Huo and Mingyu Ding and Nanyi Fei and Zhiwu Lu},
journal = {Machine Intelligence Research},
title = {Cross-modal Contrastive Learning for Generalizable and Efficient Image-text Retrieval},
year = {2023},
volume = {20},
issue = {4},
pages = {569-582},
doi = {10.1007/s11633-022-1386-4}
}

@Article {MIR-2022-06-193,
author = {Fei-Long Chen and Du-Zhen Zhang and Ming-Lun Han and Xiu-Yi Chen and Jing Shi and Shuang Xu and Bo Xu},
journal = {Machine Intelligence Research},
title = {VLP: A Survey on Vision-language Pre-training},
year = {2023},
volume = {20},
issue = {1},
pages = {38-56},
doi = {10.1007/s11633-022-1369-5}
}

@article{liu2024rgbt,
  title={Rgbt tracking via challenge-based appearance disentanglement and interaction},
  author={Liu, Lei and Li, Chenglong and Xiao, Yun and Ruan, Rui and Fan, Minghao},
  journal={IEEE Transactions on Image Processing},
  volume={33},
  pages={1753--1767},
  year={2024},
  publisher={IEEE}
}
